\documentclass[sigconf]{acmart}
\usepackage{multirow}
\AtBeginDocument{%
  }

\copyrightyear{2026}
\acmYear{2026}
\setcopyright{cc}
\setcctype{by}
\acmConference[MM '26] {Proceedings of the 34th ACM International Conference on Multimedia}{November 10--14, 2026}{Rio de Janeiro, Brazil}
\acmBooktitle{Proceedings of the 34th ACM International Conference on Multimedia (MM '26), November 10--14, 2026, Rio de Janeiro, Brazil}
\acmDOI{10.1145/3767308.3834902}
\acmISBN{979-8-4007-2213-4/2026/11}

\begin{document}

\title{DAPointRWKV: Domain Adaptive Vision RWKV for Point Cloud Completion}

\author{Yinghui Li}
\orcid{0009-0002-8838-969X}
\affiliation{%
  \institution{Deakin University}
  \city{Waurn Ponds}
  \state{Victoria}
  \country{Australia}
}
\email{s218161821@deakin.edu.au}

\author{Qianyu Zhou}
\orcid{0000-0002-5331-050X}
\correspondingauthor
\affiliation{%
  \institution{The University of Tokyo}
  \city{Tokyo}
  \country{Japan}}
\email{q-zhou@mi.t.u-tokyo.ac.jp}

\author{Hao Yang}
\orcid{0009-0007-0032-4186}
\affiliation{%
  \institution{Shanghai Jiao Tong University}
  \city{Shanghai}
  \country{China}
}
\email{yanghao.cs@sjtu.edu.cn}

\author{Richard Dazeley}
\orcid{0000-0002-6199-9685}
\affiliation{%
 \institution{Deakin University}
 \city{Waurn Ponds}
 \state{Victoria}
 \country{Australia}
}
\email{richard.dazeley@deakin.edu.au}

\author{Ye Zhu}
\orcid{0000-0003-4776-4932}
\affiliation{%
  \institution{Deakin University}
  \city{Burwood}
  \state{Victoria}
  \country{Australia}
}
\email{ye.zhu@deakin.edu.au}

\author{Xuequan Lu}
\orcid{0000-0003-0959-408X}
\correspondingauthor
\affiliation{%
  \institution{The University of Western Australia}
  \city{Perth}
  \state{Western Australia}
  \country{Australia}}
\email{bruce.lu@uwa.edu.au}

\renewcommand{\shortauthors}{Yinghui Li et al.}

\begin{abstract}
Domain adaptive point cloud completion (DA PCC) aims to bridge the geometric discrepancy between labeled source domains and unlabeled target domains for accurate 3D shape completion. Existing methods based on CNNs, Transformers, and Mamba-like sequence models still face challenges from limited receptive fields, high computational complexity, insufficient geometric bias, and serialization sensitivity. Recently, Receptance Weighted Key Value (RWKV) provides a linear-complexity alternative for long-range sequence modeling. However, adapting RWKV to DA PCC remains challenging due to the lack of explicit geometric priors, the mismatch between recurrent modeling and coarse-to-fine generation, and unstable target-domain adaptation. To address these challenges, we propose DAPointRWKV, a novel RWKV-based framework for DA PCC. DAPointRWKV consists of three components: Geometry-Aware RWKV (GA-RWKV) Backbone, RWKV-Oriented Progressive Decoder (RWKV-PD), and Dual Structural Regularization (DSR). Specifically, GA-RWKV introduces geometry-aware recurrent representation learning, RWKV-PD bridges RWKV token representations with coarse-to-fine reconstruction through local token-guided refinement, and DSR stabilizes target-domain transfer through visible-structure anchoring and complete-shape hypothesis stabilization. Extensive experiments on synthetic and real-world benchmarks demonstrate that DAPointRWKV consistently outperforms state-of-the-art methods. 
\end{abstract}



\begin{teaserfigure}
  \includegraphics[width=\textwidth]{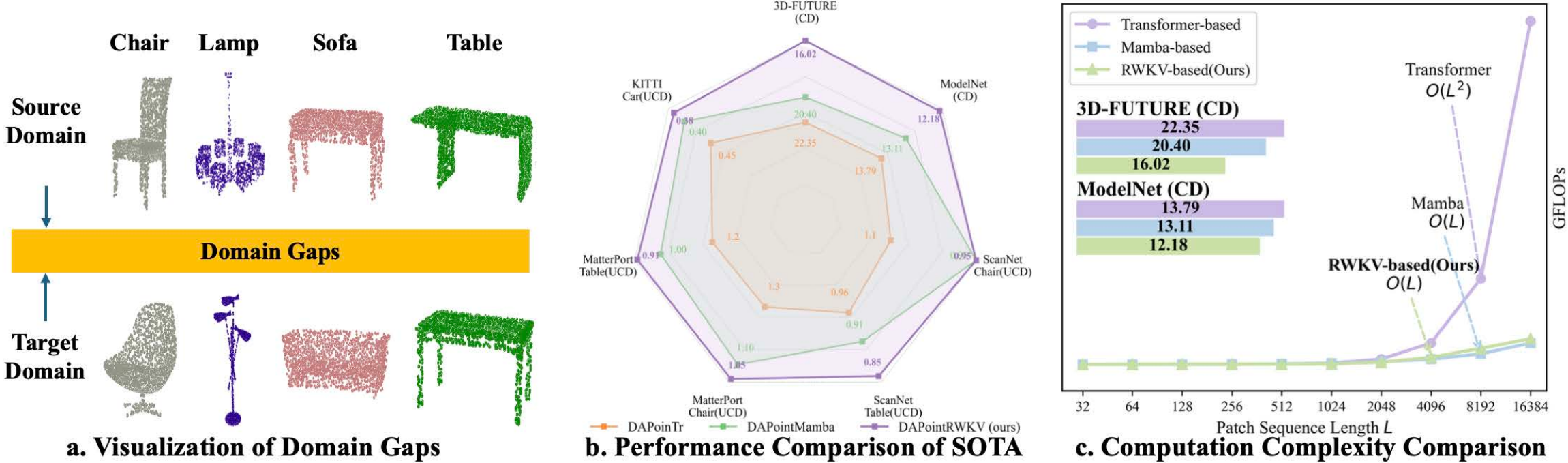}
  \caption{(a) Visualization of the domain gaps among objects of the same category from different domains, where significant variations in topology exist. (b) Compared to DAPoinTr~\cite{li2025dapointr} and DAPointMamba~\cite{li2025dapointmamba}, DAPointRWKV achieves state-of-the-art performance on challenging benchmarks. (c) Architecture trade-off in UDA Point Cloud Completion.}
  \label{fig:teaser}
\end{teaserfigure}

\maketitle

\section{Introduction}
Point cloud completion (PCC) aims to predict complete 3D shapes from partial observations and plays a fundamental role in multimedia applications involving 3D content understanding~\cite{sun2024review, chen2023clip2scene}, immersive interaction~\cite{bruder2014poster,sun2023dynamic}, autonomous driving~\cite{yang2024visual,yue2018lidar,zeng2018rt3d}, and scene understanding~\cite{jaritz2019multi, zheng2013beyond,dai2018scancomplete, gong2021omni}.
Recent supervised methods~\cite{yu2021pointr, yan2022fbnet, Chen_2023_CVPR, wang2024pointattn, yuan2018pcn,zhang2021view,wen2021pmp, lin2023hyperbolic,li2023proxyformer,fang2026point} have achieved substantial progress in recovering complete shapes from partial point clouds. However, these methods rely heavily on paired partial-complete training data, which are difficult to obtain in real-world scenarios. In practice, clean and complete point clouds are inherently difficult to acquire due to scanner limitations and occlusions~\cite{yuan2018pcn,fei2022comprehensive,wang2021unsupervised}. Moreover, their performance often drops significantly when directly transferred from synthetic to real-world scans due to the substantial domain gaps (as shown in Figure~\ref{fig:teaser}(a)) between the source and target domains. 

To address this issue, unsupervised domain adaptation (UDA) techniques~\cite{ganin2015unsupervised, kang2019contrastive,fang2024source, wilson2020survey, long2016unsupervised,pinheiro2018unsupervised,zhou2022context,zhou2022uncertainty,zhou2022generative,zhou2022domain,gu2021pit,jiang2024pcotta} have been introduced into point cloud completion to narrow the domain shifts between the labeled source domain and the unlabeled target domain. Existing UDA PCC methods can be grouped into several categories, including adversarial learning~\cite{zhang2021unsupervised, wu2025unsupervised, chen2019unpaired,xie2021style}, data reconstruction~\cite{wen2021cycle4completion}, disentangled representation learning~\cite{gong2022optimization}, and self-supervised learning~\cite{chen2019unpaired, hong2023acl, mittal2021self,yang2024syn,liu2024cloudmix}.  More recently, DAPoinTr~\cite{li2025dapointr} introduces the Transformer-based architecture into UDA PCC to align feature distributions at both local and global levels. However, as the number of point tokens increases, the quadratic complexity of attention yields prohibitive computational and memory costs, as shown in Figure~\ref{fig:teaser}(c). To improve the efficiency, DAPointMamba~\cite{li2025dapointmamba} further demonstrates that linear-complexity sequence modeling is a promising direction for UDA PCC. Despite this progress, Mamba-based models still remain limited by weak explicit geometric bias and sensitivity to point serialization. In particular, when sparse and unordered point clouds are serialized into 1D sequences, the original spatial topology is disrupted, and fine-grained local geometric cues are lost, thereby hindering effective spatial representation learning.

Meanwhile, Receptance Weighted Key Value (RWKV)~\cite{peng2023rwkv}, another efficient recurrent architecture, models long-range dependencies with linear computational complexity through weighted recurrent key-value aggregation. Recent studies~\cite{duan2024vision, he2025pointrwkv, sun2025rwkv3d,rao2026serialized,wang2025rs,fei2024diffusion,dai2025stylerwkv} have demonstrated the effectiveness of RWKV-based methods for both vision and point cloud analysis tasks. Vision-RWKV~\cite{duan2024vision} validates RWKV as an efficient visual backbone with favorable scalability and long-range modeling capability, while PointRWKV~\cite{he2025pointrwkv}, RWKV3D~\cite{sun2025rwkv3d}, and Serialized PointRWKV~\cite{rao2026serialized} further verify its strong performance and efficiency in 3D point cloud analysis. Despite its gratifying progress, directly applying existing RWKV-based approaches to UDA PCC remains non-trivial due to the lack of consideration of domain shifts and tailored designs. Therefore, improving the transferability of RWKV-based models for UDA PCC remains an underexplored problem.

In our work, we aim to enhance the transferability of RWKV-like models in UDA PCC. Our motivations mainly lie in three aspects. \textit{Firstly}, although RWKV offers efficient global context aggregation, it lacks explicit local geometric inductive bias, making it difficult to capture fine-grained structural priors in irregular point clouds. This issue is particularly critical for UDA PCC, where reliable recovery of missing regions requires strong local geometric awareness and structural continuity under domain shift.
\textit{Secondly}, PCC is inherently a coarse-to-fine geometric generation problem, whereas RWKV is originally more suitable for sequential representation learning. As a result, directly replacing an existing encoder with RWKV cannot naturally establish the connection between recurrent token representations and progressive point-wise reconstruction. \textit{Finally}, RWKV’s recurrent weighted aggregation can become unstable under unlabeled target-domain adaptation. Real-world scans often contain noise, irregular missing patterns, and domain-specific clutter, which may bias recurrent feature propagation and lead to unreliable completion hypotheses.

In this paper, we propose DAPointRWKV, the first RWKV-based UDA framework explicitly tailored for domain adaptive point cloud completion. As presented in Figure~\ref{fig:teaser}(b), our method offers strong transferability to the target domain, linear computational complexity, and effective modeling of long-range dependencies. DAPointRWKV comprises three key components. \textit{Firstly}, the Geometry-Aware RWKV (GA-RWKV) Backbone is developed to enhance RWKV with explicit local geometric priors, enabling geometry-aware recurrent token modeling for irregular point clouds. \textit{Secondly}, we design an RWKV-Oriented Progressive Decoder (RWKV-PD) to bridge RWKV token representations and coarse-to-fine geometric reconstruction through global shape semantics and local token-guided refinement. \textit{Finally}, we introduce a Dual Structural Regularization (DSR) module to stabilize target-domain adaptation, where Observation-Anchor Constraint (OAC) enforces visible-structure anchoring, and Hypothesis-Stabilizing Constraint (HSC) regularizes the stability of complete-shape hypotheses under re-observation. Extensive experiments on both synthetic and real-world benchmarks demonstrate the effectiveness of the proposed framework. Our DAPointRWKV consistently outperforms existing state-of-the-art UDA PCC methods, verifying the advantage of geometry-aware recurrent modeling for cross-domain point cloud completion. 
Furthermore, these results highlight the potential of recurrent sequence modeling as an efficient alternative for addressing domain shifts in irregular 3D data, providing a new perspective for future cross-domain point cloud understanding.

In summary, our contributions are threefold:
\begin{itemize}
\item We propose DAPointRWKV, a novel RWKV-based framework for point cloud completion. To the best of our knowledge, this is the first work that explores the transferability of RWKV-based models in DA PCC.

\item We introduce three tailored designs for RWKV-based DA PCC, including Geometry-Aware RWKV Backbone to enhance recurrent token mixing with explicit local geometric priors, RWKV-Oriented Progressive Decoder to bridge RWKV representations and coarse-to-fine point generation, and Dual Structural Regularization to stabilize unlabeled target-domain adaptation.

\item
Extensive experiments on both synthetic and real-world datasets demonstrate that our DAPointRWKV consistently outperforms existing state-of-the-art methods for domain adaptive point cloud completion.
\end{itemize}

\begin{figure*}[t]
    \centering
    \includegraphics[width=1.0\linewidth]{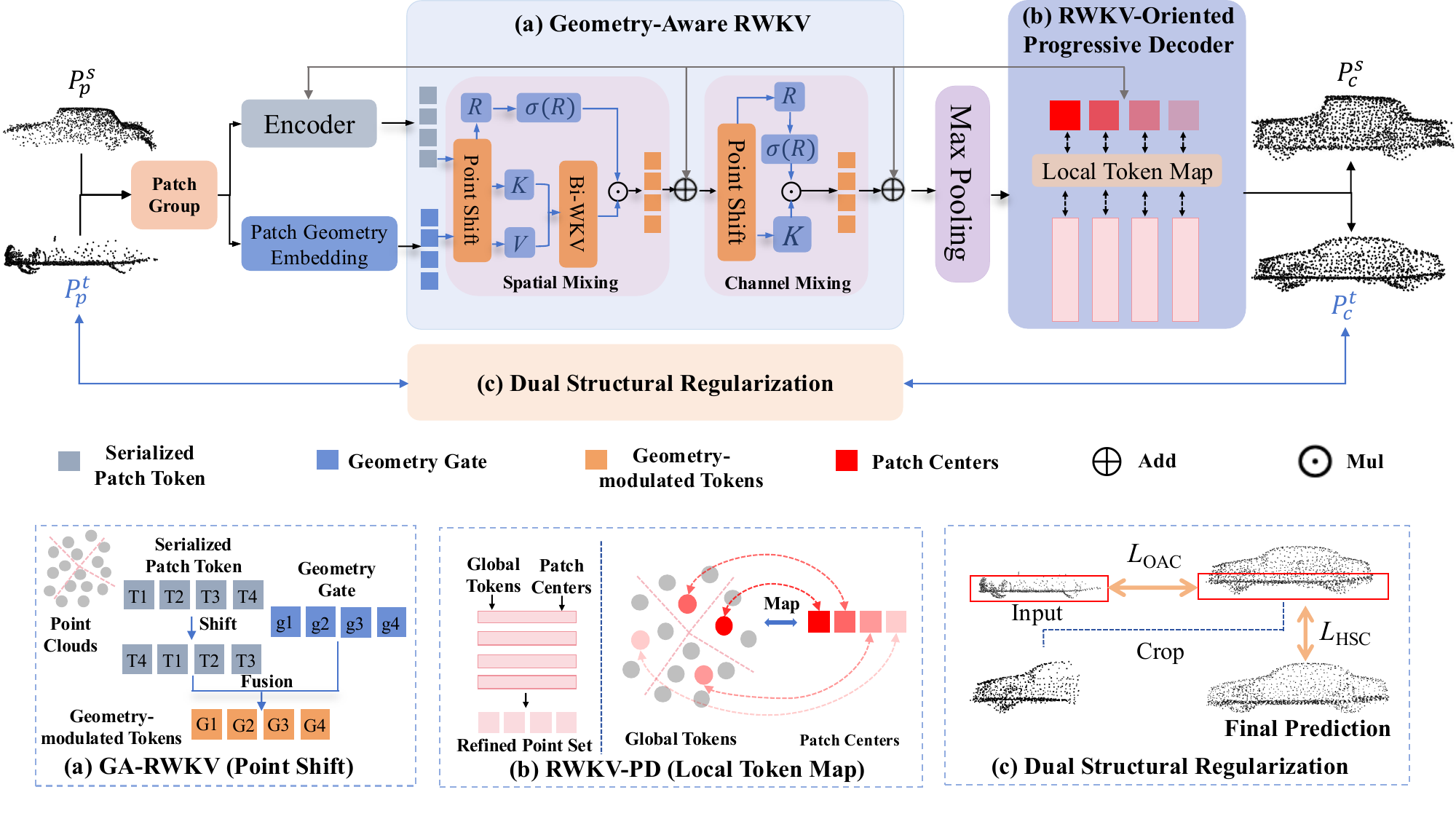}
    \caption{The framework of DAPointRWKV for cross-domain point cloud completion, including three key components: (a) Geometry-Aware RWKV Backbone is designed to inject local-level geometric priors into RWKV recurrent token mixing for irregular point clouds. (b) The RWKV-Oriented Progressive Decoder is proposed to convert RWKV features into progressively refined point sets through coarse-to-fine reconstruction with local token-guided refinement. (c) Dual Structural Regularization is presented to regularize target-domain completion by enforcing observation anchoring and hypothesis stabilization.} 
    \label{framework}
\end{figure*}

\section{Related Work}
\subsection{Domain Adaptive Point Cloud Completion}

Point cloud completion (PCC) aims to recover complete shapes from partial inputs. Existing supervised PCC methods~\cite{yuan2018pcn,huang2020pf,xie2020grnet, yu2021pointr, wang2024pointattn,xiang2021snowflakenet,tang2022lake, yan2022shapeformer,mittal2022autosdf,groueix2018papier,tchapmi2019topnet,chang2021finerpcn, li2025genpc,yan2025symmcompletion,fang2025multi} have achieved remarkable progress by learning from paired partial-complete point cloud data. However, their reliance on paired annotations limits their applicability in real-world scenarios, where complete ground truth is difficult to obtain, and substantial domain gaps often exist between synthetic and real scans. To alleviate this issue, unsupervised domain adaptation (UDA) techniques have been introduced to PCC. Pcl2pcl~\cite{chen2019unpaired} initially proposes to align partial and complete latent spaces through dual auto-encoders and adversarial mapping, while ShapeInversion~\cite{zhang2021unsupervised} leverages a pretrained GAN prior and performs completion through latent inversion. Based on Pcl2pcl, Cycle4Completion~\cite{wen2021cycle4completion} introduces bidirectional cycle transformations between partial and completion latent spaces to improve geometric correspondence. Furthermore, OptDE~\cite{gong2022optimization} explicitly disentangles occlusion, domain, and shape factors to address both input-domain and output-domain gaps. ACL-SPC~\cite{hong2023acl} further improves target-domain generalization through adaptive closed-loop completion. More recently, DAPoinTr~\cite{li2025dapointr} introduces Transformer-based sequence-wise features alignment at both the global and local levels into UDA PCC, while DAPointMamba~\cite{li2025dapointmamba} shows that linear-complexity sequence modeling~\cite{zhang2025point,yang2025pointdgmamba} is also applicable to this task. However, existing UDA PCC methods still either suffer from limited receptive fields or high computational cost, motivating the exploration of more efficient transferable architectures.

\subsection{Receptance Weighted Key Value (RWKV)}
RWKV~\cite{peng2023rwkv,peng2024eagle, yang2026beyond, yang2025restore, zhang2026rwkv,xu2025rwkvquant,pan2025exploring} has recently emerged as an efficient sequence modeling architecture, offering global receptive fields and strong long-range dependency modeling with linear computational complexity. Its effectiveness primarily stems from the linear WKV attention mechanism and the spatial/channel mixing design, which balance local feature interactions and global context propagation through gated recurrent computation. Such a design provides efficient parallel training and low-cost inference, making it particularly appealing for large-scale point cloud data. Vision-RWKV~\cite{duan2024vision} demonstrates its effectiveness as a scalable visual backbone with strong scalability and favorable efficiency-performance trade-offs. Building on this, PointRWKV~\cite{he2025pointrwkv} first introduces this architecture to point cloud analysis and combines global recurrent modeling with local feature aggregation. Subsequent works further adapt it to point clouds from different perspectives. RWKV3D~\cite{sun2025rwkv3d} improves local geometric modeling through redesigned feature mixing and multi-head shift strategies, while Serialized PointRWKV~\cite{rao2026serialized} highlights the importance of point serialization and reordering for sequential point modeling. More recently, PointDGRWKV~\cite{yang2026pointdgrwkv} extends it to domain generalization for point cloud classification by improving geometry perception and distribution robustness, while RWKV-PCSSC~\cite{he2025rwkv} explores RWKV for point cloud semantic scene completion by redesigning the seed generator and multiple deconvolution. 
These studies consistently demonstrate the potential of RWKV for efficient point cloud representation learning and show that RWKV-style recurrent modeling can be effectively adapted to diverse 3D tasks.
Nevertheless, existing RWKV-based point methods mainly focus on analysis tasks or semantic scene completion, rather than UDA PCC. More importantly, they do not explicitly consider domain shifts and task-specific designs for completion. To the best of our knowledge, this is the first work that investigates the transferability of the RWKV-based model for UDA PCC.

\section{Methodology}
In this section, we introduce DAPointRWKV, a novel RWKV-based framework for domain adaptive point cloud completion (DA PCC). The proposed framework is motivated by three key challenges in extending RWKV to DA PCC. Firstly, the recurrent token mixing mechanism in vanilla RWKV lacks explicit geometric awareness, limiting its ability to capture local structural priors in irregular point clouds. Secondly, RWKV-like sequential representation learning is not naturally aligned with the coarse-to-fine geometric generation process required for PCC. Finally, the absence of target-domain annotations makes cross-domain adaptation inherently unstable, often leading to unreliable optimization and degraded performance. 

To address these challenges, DAPointRWKV, as illustrated in Figure~\ref{framework}, consists of three components: Geometry-Aware RWKV (GA-RWKV) Backbone, RWKV-Oriented Progressive Decoder (RWKV-PD), and Dual Structural Regularization (DSR). Specifically, GA-RWKV enhances RWKV with geometry-aware recurrent representation learning to better capture local structural information in irregular point clouds; RWKV-PD bridges token representations with progressive geometric reconstruction by combining global shape semantics and local token-guided refinement; and DSR serves as the adaptation module to stabilize unlabeled target-domain learning through observation anchoring and hypothesis stabilization. By integrating these components into a unified framework, DAPointRWKV achieves robust cross-domain feature learning and accurate point cloud completion.

\subsection{Geometry-Aware RWKV Backbone} 
RWKV~\cite{peng2023rwkv,wang2025rs, fei2024diffusion, peng2025rwkv, hou2024rwkv} provides an attractive alternative to Transformer-based architectures by enabling efficient long-range dependency modeling with linear-complexity recurrent computation. This property is particularly desirable for PCC, where understanding missing parts requires both global context aggregation and effective feature interaction across spatial regions.
However, directly applying RWKV to serialized point tokens leads to weak local geometric awareness, since spatially neighboring points may be far apart in sequence space. Therefore, explicitly injecting local-level geometric priors into recurrent token mixing is necessary to preserve local structural consistency. To address this issue, we propose a Geometry-Aware RWKV (GA-RWKV) Backbone that introduces geometric awareness into RWKV and improves its ability to model local structural details in irregular point clouds.

Given an input partial point cloud $P_{part} \in \mathbb{R}^{B \times N \times 3}$, we first convert it into a sequence of patch tokens. Specifically, following the serialized grouping strategy~\cite{li2025dapointmamba}, the input is partitioned into $M$ local patches, and each patch is embedded as a latent token together with its positional embedding, yielding the input token sequence $X^0 \in \mathbb{R}^{B \times M \times C}$. To further encode local structural information, we construct a geometry embedding for each patch based on the relative coordinates of points with respect to the patch center and their corresponding point-to-center distances. Formally, for the $i$-th patch $\mathcal{G}_i$, the geometry embedding is defined as:
\begin{equation}
g_i = \operatorname{Mean}\!\left(\phi\left([\,p - c_i,\ \|p - c_i\|_2\,]\right)\right), \quad p \in \mathcal{G}_i
\end{equation}
where $c_i$ denotes the patch center and $\phi(\cdot)$ is a learnable nonlinear projection. Collecting all patch-wise geometry embeddings yields the geometry embedding sequence $G \in \mathbb{R}^{B \times M \times C}$. In this way, each patch token is associated with an explicit local geometry descriptor that provides structural cues for subsequent recurrent modeling.

Built upon the token sequence $X^0$ and the corresponding geometry embedding sequence $G$, GA-RWKV performs geometry-aware feature learning through stacked RWKV blocks. Unlike vanilla RWKV, which primarily models token interactions via sequential dependency modeling, our GA-RWKV injects local-level geometric information into the spatial mixing process. Specifically, given the token sequence $X^l$ at the $l$-th layer, we first apply a point-sequence shift operation to obtain a shifted representation:
\begin{equation}
X_{\mathrm{shift}}^{l} = \mathrm{Shift}(X^l)
\end{equation}
Then, a geometry-aware gate is generated from $G$ to adaptively modulate the interaction between the original and shifted sequences:
\begin{equation}
A = \lambda \cdot \sigma\!\left(\psi(G)\right),
\qquad
\tilde{X}^{\,l} = X^{l} + A \odot \left(X_{\mathrm{shift}}^{\,l} - X^{l}\right)
\end{equation}
where $\psi(\cdot)$ denotes an MLP, $\sigma(\cdot)$ is the sigmoid function, and $\lambda$ is a learnable scale parameter. The resulting sequence $\tilde{X}^{\,l}$ is used in the GA-RWKV spatial mixing branch, enabling recurrent token interaction to be conditioned on both sequential context and local geometric structure. Formally, the $l$-th GA-RWKV block can be represented as:
\begin{equation}
\begin{aligned}
\hat{X}^{\,l} &= X^{l} + \mathrm{SpatialMix}\!\left(\mathrm{LN}(\tilde{X}^{l}),\, G\right), \\
X^{l+1} &= \hat{X}^{\,l} + \mathrm{ChannelMix}\!\left(\mathrm{LN}(\hat{X}^{\,l})\right)
\end{aligned}
\end{equation}
where $X^l \in \mathbb{R}^{B \times M \times C}$ denotes the token sequence at the $l$-th layer. Through stacked Geometry-Aware RWKV blocks, the backbone captures both long-range dependencies and local structural priors in irregular point clouds. The output tokens are then aggregated into a global feature for completion, while the patch-wise features are retained to guide the subsequent progressive decoder.

\subsection{RWKV-Oriented Progressive Decoder}
Although existing coarse-to-fine methods~\cite{xiang2021snowflakenet,li2025dapointr,li2025dapointmamba, zhao2022pcunet,chang2021finerpcn} progressively refine point clouds in a hierarchical manner, they do not explicitly incorporate patch-wise structural tokens into local geometric refinement. Consequently, the structural cues encoded by GA-RWKV cannot be fully exploited during progressive reconstruction. To address this limitation, we design an RWKV-Oriented Progressive Decoder that transforms the patch-wise token representations learned by GA-RWKV into progressively refined point sets via global shape semantics and local token-guided refinement.

Given the encoded token sequence $X^L$ from GA-RWKV, we aggregate it into a global feature $f_g$, from which a coarse point cloud $P_{coarse}$ is generated for subsequent progressive refinement:
\begin{equation}
P_{coarse} = \mathcal{H}_{\mathrm{seed}}(f_g)
\end{equation}

To preserve the observed geometry, we concatenate $P_{coarse}$ with the input partial point cloud $P_{part}$ and then apply farthest point sampling (FPS) to obtain the initial point set for refinement:
\begin{equation}
P_{init} = \operatorname{FPS}\!\left([P_{coarse}; P_{part}],\, N_{init}\right)
\end{equation}
where $[\cdot;\cdot]$ denotes point-wise concatenation and $N_{init}$ is the number of sampled points.
Starting from $P_{init}$, the decoder progressively refines the reconstruction through multiple stacked refinement stages. We set $P_0 = P_{init}$. At the $k$-th stage, the point set $P_k$ is updated as:
\begin{equation}
P_k = \mathcal{D}_k(P_{k-1}, f_g, C, T)
\end{equation}
where $\mathcal{D}_k(\cdot)$ denotes the $k$-th refinement stage, $C=\{c_i\}_{i=1}^{M}$ denotes the patch centers, and $T=\{t_i\}_{i=1}^{M}$ denotes the patch-wise token features in $X^L$. Specifically, for each point in the current point set, we retrieve its nearest patch center from $C$ and use the corresponding token feature in $T$ as local structural guidance. In this way, each refinement stage is conditioned on both the global RWKV feature and local-level RWKV representations.

At each stage, the decoder predicts coordinate offsets for the upsampled point set, and the refined point cloud is updated in a coarse-to-fine manner:
\begin{equation}
P_k = \operatorname{Upsample}(P_{k-1}) + \Delta P_k
\end{equation}
where $\Delta P_k$ denotes the predicted offset at the $k$-th stage. Through multiple stacked refinement stages, the RWKV-PD progressively transforms the coarse prediction into a dense and geometrically consistent complete point cloud. Overall, the proposed RWKV-PD bridges RWKV-based sequential representation learning and progressive geometric reconstruction by combining global shape semantics with local token-guided refinement.

\subsection{Dual Structural Regularization}
Although the geometry-aware backbone and RWKV-Oriented Progressive Decoder improve structural modeling and progressive reconstruction, they do not explicitly address the domain shifts between the labeled source domain and the unlabeled target domain. To enable effective cross-domain adaptation, we further introduce the \textbf{Dual Structural Regularization (DSR)} module, which consists of \textbf{Observation-Anchor Constraint (OAC)} and \textbf{Hypothesis-Stabilizing Constraint (HSC)}. These two constraints regularize target-domain completion from two complementary perspectives, namely visible-structure anchoring and complete-shape hypothesis stabilization.

\noindent\textbf{Observation-Anchor Constraint.}
Given an unlabeled target-domain partial point cloud $P_p^t$ and its predicted completion $\hat{P}^t$, OAC constrains the prediction to remain compatible with the reliable observed geometry. Specifically, we impose a one-way point-to-shape constraint from the target partial input to the predicted complete shape:
\begin{equation}
\mathcal{L}_{\mathrm{OAC}}
=
\frac{1}{|P_p^t|}
\sum_{x \in P_p^t}
\min_{y \in \hat{P}^t} \|x-y\|_2^2
\end{equation}
Unlike a symmetric reconstruction loss, this formulation does not force the prediction to collapse toward the visible subset; therefore, it preserves the freedom to generate missing structures. In this way, OAC serves as a visible-structure anchor for RWKV-based target-domain completion.

\noindent\textbf{Hypothesis-Stabilizing Constraint.}
While OAC anchors the prediction to the current observation, it does not explicitly regularize whether the predicted complete shape itself is stable under varying partial views. To address this issue, we introduce HSC. Specifically, we treat the target prediction $\hat{P}^t$ as a stop-gradient teacher shape, denoted by $\tilde{P}^t$, randomly crop it to simulate another partial observation, and feed the re-cropped point cloud back into the model to obtain a new prediction $\hat{P}_r^{t}$. We then enforce structural consistency between the two completed shapes:
\begin{equation}
\mathcal{L}_{\mathrm{HSC}}
= \frac{1}{2}\left(
\frac{1}{|\hat{P}_r^{t}|}
\sum_{x \in \hat{P}_r^{t}}
\min_{y \in \tilde{P}^t} \|x-y\|_2
+
\frac{1}{|\tilde{P}^t|}
\sum_{y \in \tilde{P}^t}
\min_{x \in \hat{P}_r^{t}} \|x-y\|_2
\right)
\end{equation}
By enforcing consistency across predictions derived from diverse partial observations of the same underlying shape, HSC regularizes the latent shape hypothesis and stabilizes the reconstruction process on unlabeled target samples, preventing observation-biased or degenerate solutions.

Overall, DSR serves as the key adaptation component of DAPointRWKV. By jointly enforcing observation anchoring and hypothesis stabilization on unlabeled target-domain samples, it effectively stabilizes cross-domain completion learning and improves target-domain completion performance.

\subsection{Training and Inference}
During the training, the source-domain partial point cloud $P_p^{s}$ is supervised by its complete ground truth $P_c^{s}$, while the unlabeled target-domain samples are regularized by the proposed Dual Structural Regularization (DSR). Accordingly, the total loss function can be formulated as:
\begin{equation}
\mathcal{L}_{\mathrm{total}} = \mathcal{L}_{\mathrm{source}} + \lambda_{\mathrm{OAC}} \mathcal{L}_{\mathrm{OAC}} + \lambda_{\mathrm{HSC}} \mathcal{L}_{\mathrm{HSC}}
\end{equation}
where $\mathcal{L}_{\mathrm{source}}$ denotes the source-domain completion loss, and $\lambda_{\mathrm{OAC}}$ and $\lambda_{\mathrm{HSC}}$ are the balancing coefficients for Observation-Anchor Constraint and Hypothesis-Stabilizing Constraint, respectively.

During the inference, only the trained GA-RWKV backbone and RWKV-PD are used to complete target-domain partial point clouds. No source-domain data or additional adaptation process is required. Given a target-domain partial input, the model directly predicts the corresponding complete point cloud, enabling efficient deployment in unseen target domains.

\section{Experiments}
We conduct extensive experiments to evaluate the effectiveness of the proposed method. In addition to quantitative comparisons on synthetic benchmark datasets and widely used real-world partial scans, we further provide qualitative visualization of completion results to demonstrate the effectiveness of our approach.

\subsection{Experiment Settings}

\noindent\textbf{Datasets.} Following the standard evaluation protocol widely adopted in UDA PCC~\cite{wen2021cycle4completion, gong2022optimization,li2025dapointr}, we use CRN~\cite{wang2020cascaded} as the labeled source dataset and evaluate on synthetic and real-world target datasets. Specifically, we use 26,863 samples from categories shared by CRN and target datasets. Synthetic targets include 3D-FUTURE~\cite{fu20213d} and ModelNet~\cite{wu20153d}. For real-world scans, we evaluate on ScanNet~\cite{dai2017scannet}, MatterPort3D~\cite{chang2017matterport3d}, and KITTI~\cite{geiger2012we}, covering indoor chair/table and outdoor car completion scenarios. Following prior practice, only shared categories between source and target domains are used for training and evaluation. All partial inputs are uniformly down-sampled to 2048 points for unpaired training and inference.

\noindent\textbf{Implementation Details.}
All experiments are conducted on an RTX 4090 with 64GB RAM. For the training, we employ an initial learning rate of $1 \times 10^{-3}$ and a weight decay of $5 \times 10^{-2}$. The batch size is set to 8. To balance losses, weights of $\mathcal{L}_{OAC}$ and $\mathcal{L}_{HSC}$ are set to 0.2 and 0.2, respectively. We build the basic network by adapting the core Vision-RWKV~\cite{duan2024vision} blocks for point cloud completion. Following previous works~\cite{li2025dapointr,wen2021cycle4completion, gong2022optimization,hong2023acl,zhang2021unsupervised}, we adopt the mean Chamfer Distance (CD) as our metric to evaluate the performance on the synthetic 3D-FUTURE and ModelNet datasets in which complete shapes are available for testing. For real-world scans without ground truth, we evaluate the completion quality using Unidirectional Chamfer Distance (UCD) and Unidirectional Hausdorff Distance (UHD) from partial input to the predicted completion. 
\begin{table}[t!]
\centering
\footnotesize
\caption{Cross-domain completion results on 3D-FUTURE.  Chamfer Distance (CD)$\downarrow$ is used as the metric, and the scale factor is $10^4$. Lower is better. }
\begin{tabular}{l|l|cccccc}
\toprule
Methods&Avg&Cabinet&Chair&Lamp&Sofa&Table \\
\midrule
Pcl2Pcl~\cite{chen2019unpaired}&92.83&57.23&43.91&157.86&63.23&141.92\\
ShapeInv.~\cite{zhang2021unsupervised}&53.21 &38.54&26.30&48.57&44.02&108.60\\
Cycle4Comp.~\cite{wen2021cycle4completion}&45.39 &32.62&34.08&77.19&43.05&40.00\\
ACL-SPC~\cite{hong2023acl}&35.97&70.12&23.87&31.75&28.74&25.38\\
OptDE~\cite{gong2022optimization}&28.99&28.37&21.87&29.92&37.98&26.81 \\
DAPoinTr~\cite{li2025dapointr}&22.35 &\textbf{18.46} &17.60 &27.91 &23.08 &24.71\\
DAPointMamba~\cite{li2025dapointmamba}&20.40&19.35&16.21&22.81&22.38&21.25\\
DAPointRWKV&\textbf{16.02}&19.01&\textbf{15.21}&\textbf{13.27}&\textbf{19.23}&\textbf{13.40}\\
\bottomrule
\end{tabular}
\label{table1}
\end{table}

\begin{table}[t!]
\caption{Cross-domain completion results on ModelNet.  Chamfer Distance (CD)$\downarrow$ is used as the metric, and the scale factor is $10^4$. Lower is better.}
\centering
\footnotesize
\setlength{\tabcolsep}{4pt}
\label{table2}
\begin{tabular}{l|l|cccccc}
\toprule
Methods&Avg&Plane&Car&Chair&Lamp&Sofa&Table \\
\midrule
Pcl2Pcl~\cite{chen2019unpaired}&68.14&18.53&17.54&43.58&126.80&38.78&163.62\\
ShapeInv.~\cite{zhang2021unsupervised}& 41.61 &3.78& 15.66& 22.25& 60.42&22.25 & 125.31\\
Cycle4Comp.~\cite{wen2021cycle4completion}& 28.65 &5.77& 11.85& 26.67& 83.34& 22.82& 21.47 \\
ACL-SPC~\cite{hong2023acl}&34.89&5.75&11.73&43.08&106.29&25.62&16.89\\
OptDE~\cite{gong2022optimization}& 15.94 &\textbf{2.18}& 9.80& 14.71& 39.74& 19.43&9.75 \\
DAPoinTr~\cite{li2025dapointr}&13.79 &2.38 &8.04 &13.83 &33.26 &12.72 &12.51\\
DAPointMamba~\cite{li2025dapointmamba}&13.11&2.30&\textbf{7.58}&13.15&32.04&12.48&11.08\\
DAPointRWKV&\textbf{12.18}&2.78&7.71&\textbf{12.20}&\textbf{29.87}&\textbf{11.84}&\textbf{8.65}\\
\bottomrule
\end{tabular}
\end{table}

\begin{table}[t!]
\caption{Cross-domain completion results on real-world scans. [UCD$\downarrow$/UHD$\downarrow$] are used as the metrics, and the scale factor is $10^4$ and $10^2$ for UCD (Unidirectional Chamfer Distance) and UHD (Unidirectional Hausdorff Distance), respectively.  A lower value of UCD or UHD is better.}
\centering
\footnotesize
\begin{tabular}{@{}l|cc|cc|c@{}}
\toprule
\multirow{2}{*}{Methods}
&\multicolumn{2}{c}{ScanNet}&\multicolumn{2}{c}{MatterPort3D}&\multicolumn{1}{c}{KITTI}\\
\cmidrule(lr){2-6}
&Chair&Table&Chair&Table&Car \\
\midrule
Pcl2Pcl~\cite{chen2019unpaired}&17.3/10.1&9.1/11.8&15.9/10.5&6.0/11.8&9.2/14.1\\
ShapeInv.~\cite{zhang2021unsupervised}&3.2/10.1 &3.3/11.9&3.6/10.0&3.1/11.8&2.9/13.8\\
Cycle4Comp.~\cite{wen2021cycle4completion}&5.1/6.4 &3.6/5.9&8.0/8.4&4.2/6.8&3.3/5.8\\
ACL-SPC~\cite{hong2023acl}&1.4/4.7& 1.8/5.1&1.8/4.8&2.1/4.9&2.0/4.9\\
OptDE~\cite{gong2022optimization}&2.6/5.5&1.9/4.6&3.0/5.5&1.9/5.3&1.6/3.5 \\
DAPoinTr~\cite{li2025dapointr}&1.1/2.7&0.96/2.7 &1.3/2.9 &1.2/\textbf{2.8}&0.45/1.8 \\
DAPointMamba~\cite{li2025dapointmamba}&0.95/2.8&0.91/2.6&1.1/3.0&1.0/3.0&0.40/2.1\\
DAPointRWKV&\textbf{0.95}/\textbf{2.7}&\textbf{0.85}/\textbf{2.6}&\textbf{1.04}/\textbf{2.9}&\textbf{0.91}/2.9&\textbf{0.38}/\textbf{1.8}\\
\bottomrule
\end{tabular}
\label{table3}
\end{table}

\begin{table*}[t] 
\caption{Ablation studies on each proposed module of our DAPointRWKV on 3D-FUTURE dataset. Chamfer Distance (CD)$\downarrow$ is used as the metric, and the scale factor is $10^4$. Lower is better. } 
\centering 
\footnotesize 
\begin{tabular}{l|cccc|ccccc|c} \toprule 

Methods&GA-RWKV&RWKV-PD&OAC&HSC&Cabinet&Chair&Lamp&Sofa&Table&Avg \\ \midrule Baseline&&&&&21.47&16.80&23.06&22.73&19.24&20.66\\ \midrule \multirow{5}{*}{DAPointRWKV} &\checkmark&&&&19.86&16.15&22.45&21.23&19.11&19.76 \\
&\checkmark&\checkmark&&&19.31&15.88&21.31&20.58&18.43&19.10 \\ &\checkmark&\checkmark&\checkmark&&19.85&16.27&25.05&23.29&20.29&20.95 \\ &\checkmark&\checkmark&&\checkmark&19.36&15.58&14.99&19.63&14.14&16.74 \\ &\checkmark&\checkmark&\checkmark&\checkmark&\textbf{19.01}&\textbf{15.21}&\textbf{13.27}&\textbf{19.23}&\textbf{13.40}&\textbf{16.02} \\ \bottomrule \end{tabular} 
\label{table4} 
\end{table*}

\subsection{Comparison Results}
We evaluate DAPointRWKV on both synthetic and real-world benchmarks to comprehensively validate its cross-domain completion capability and effectiveness under domain shift.

\noindent \textbf{Comparison Methods.}
To comprehensively evaluate the effectiveness of DAPointRWKV, we compare our method with state-of-the-art approaches in UDA PCC, including Pcl2Pcl~\cite{chen2019unpaired}, ShapeInversion~\cite{zhang2021unsupervised}, Cycle4Completion~\cite{wen2021cycle4completion}, ACL-SPC~\cite{hong2023acl}, OptDE~\cite{gong2022optimization}, DAPoinTr~\cite{li2025dapointr}, and DAPointMamba~\cite{li2025dapointmamba}. Due to the lack of publicly available training code, RWKV3D~\cite{sun2025rwkv3d}, and Serialized PointRWKV~\cite{rao2026serialized} are excluded from comparison.

\noindent \textbf{Results on 3D-FUTURE dataset.} 
Table~\ref{table1} shows the cross-domain completion results on the 3D-FUTURE benchmark. Our DAPointRWKV achieves the best overall performance, reducing the average CD from 20.40 to 16.02, compared to the 2nd best method, DAPointMamba. Additionally, unlike previous UDA PCC methods, which often struggle to bridge domain gaps, especially in complex categories such as Lamp and Table, our experimental results indicate substantial enhancements in these categories as evidenced by the significant margins of improvement (see the 5th and 7th columns in Table~\ref{table1}). This is because our method can better tackle domain discrepancies and is particularly effective when handling categories with significant geometric variation and more complex missing regions. Although the result on the Cabinet is slightly lower than DAPoinTr, the overall performance proves that our model can better narrow the domain discrepancies between the source and target domains, while effectively leveraging the capacity of RWKV to capture geometric structures.

\noindent \textbf{Results on ModelNet dataset.} 
As shown in Table~\ref{table2}, DAPointRWKV demonstrates superior average performance compared to the state-of-the-art methods. Our method consistently improves completion quality across most categories, further demonstrating the strong cross-domain adaptation capabilities of the proposed modules on synthetic datasets. Specifically, our model consistently achieves remarkable improvement in categories Lamp and Table (see the 6th and 8th columns in Table~\ref{table2}). This superiority can be attributed to our innovative designs, which effectively alleviate domain shifts and close domain gaps at the object level. This remarkable improvement underscores the robustness and adaptability of the innovative design elements incorporated within our model.

\noindent \textbf{Results on Real-World Scans.} 
Table~\ref{table3} reports quantitative results on three widely used real-world benchmarks: ScanNet, MatterPort3D, and KITTI. We evaluate completion quality using Unidirectional Chamfer Distance (UCD) and Unidirectional Hausdorff Distance (UHD), which measure overall geometric fidelity and worst-case point-wise deviation, respectively. As shown in Table~\ref{table3}, DAPointRWKV achieves the best or competitive performance across most real-world benchmarks, demonstrating its ability to better preserve both local details and global shape structure under real-world domain shifts. In particular, the gains are especially notable for the Table category of ScanNet and MatterPort3D datasets, where the inputs often exhibit severe incompleteness and irregular missing patterns. On KITTI, our method consistently achieves the best performance on Car, highlighting its effectiveness on sparse outdoor scans with limited visible geometry. These results further confirm that the proposed adaptation strategy can effectively stabilize target-domain completion and produce more reliable and geometrically coherent predictions in practical cross-domain scenarios.

\begin{figure*}[t]
\centering
\includegraphics[width=1.0\textwidth]{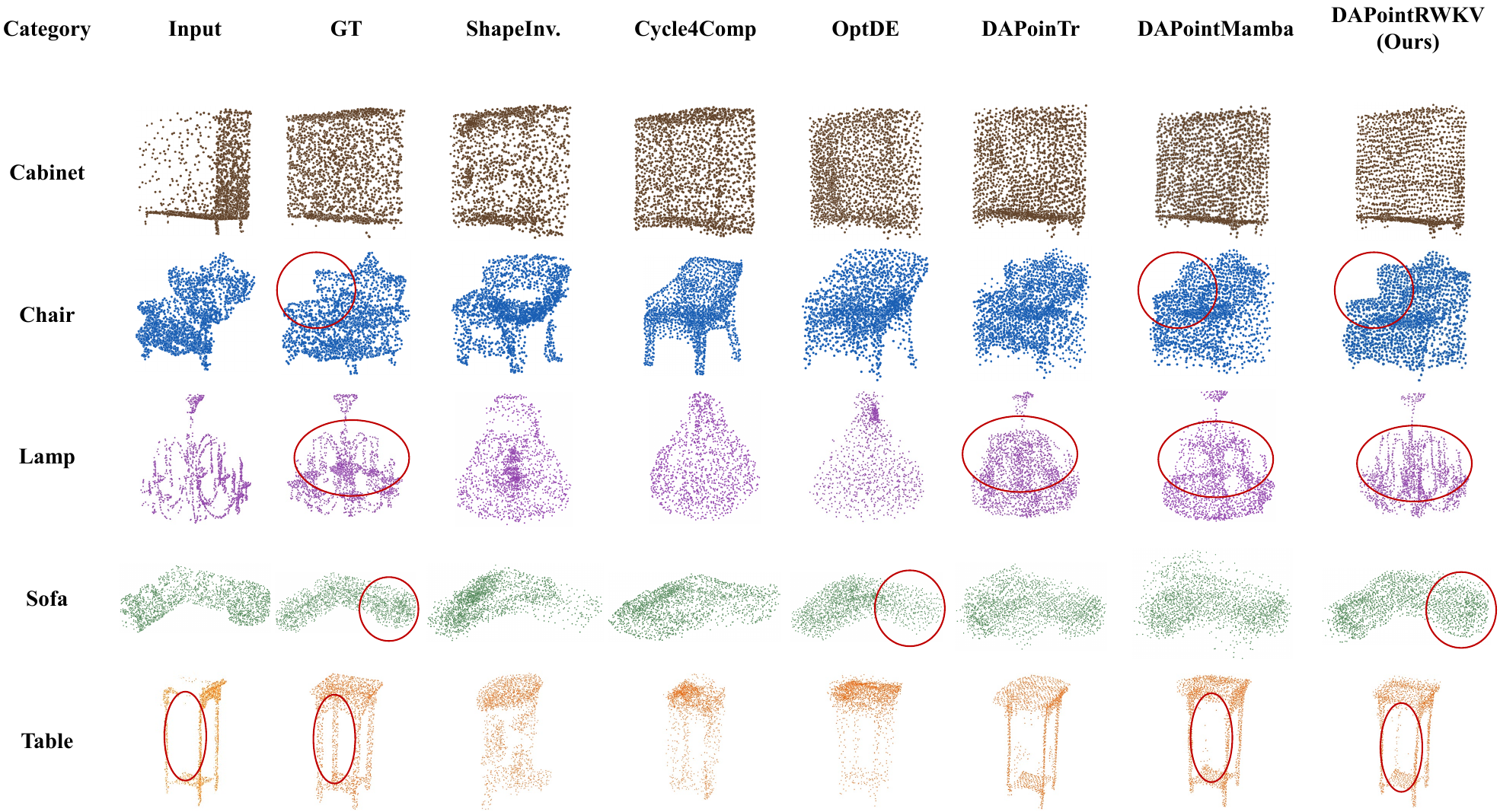} 
\caption{Visualization comparisons between ours and state-of-the-art UDA PCC methods on 3D-FUTURE dataset. }
\label{3D-FUTURE-Visualization}
\end{figure*}

\subsection{Ablation Studies}
In this section, we conduct ablation studies on the 3D-FUTURE benchmark to evaluate the contribution of each proposed component, including the Geometry-Aware RWKV (GA-RWKV) Backbone, RWKV-Oriented Progressive Decoder (RWKV-PD), and Dual Structural Regularization (DSR). Following the standard UDA PCC setting, we use the CRN dataset as the source domain and 3D-FUTURE as the target domain.

\noindent \textbf{Effectiveness of each component.}
We first constructed a basic point cloud completion network by adapting vision-RWKV~\cite{duan2024vision} blocks (denoted as \textbf{Baseline}) to train on the source domain and directly tested it on the target domain. As shown in Table~\ref{table4}, the basic PCC model achieves an average CD of 20.66, serving as a lower-bound reference for RWKV-based cross-domain completion. After incorporating GA-RWKV, the performance is consistently improved across all categories, reducing the average CD to 19.76. The improvement can be attributed to the geometry-aware recurrent modeling strategy, which strengthens RWKV token interactions with local structural cues, thereby enhancing geometric perception in irregular point clouds. The gains are more pronounced on Lamp, Sofa, and Table, indicating that local geometry-aware recurrent modeling is especially helpful for categories with more complex structures and missing patterns. 
Building upon GA-RWKV, we further introduce RWKV-PD, which explicitly couples RWKV token representations with progressive point generation through local token-guided refinement. This further reduces the average CD from 19.76 to 19.10. The improvement is particularly evident in categories with more intricate local structures, such as Lamp (from 22.45 to 21.31), demonstrating that RWKV-PD is effective in translating recurrent token features into finer point-wise geometric recovery. 

Finally, we investigate the contributions of the two components of DSR (OAC and HSC). Directly applying OAC alone leads to significant degradation, increasing the average CD from 19.10 to 20.95, which is even worse than the baseline. This result indicates that visible-structure anchoring alone cannot provide sufficiently reliable target-domain regularization, because constraining predictions to the observed partial input tends to bias the model toward conservative completion and suppress recovery of missing structures. By contrast, using HSC alone reduces the average CD to 16.74, with especially large gains on Lamp (from 21.31 to 14.99) and Table (from 18.43 to 14.14), demonstrating the effectiveness of complete-shape hypothesis stabilization across varying partial views. When OAC and HSC are jointly integrated as DSR, the average CD is further reduced to 16.02, yielding the best overall performance. 

These results show that stable cross-domain point cloud completion requires both visible-structure anchoring and complete-shape hypothesis stabilization. Specifically, OAC regularizes the prediction with reliable observed geometry, while HSC stabilizes the complete-shape hypothesis across varying partial views. Their joint use therefore enables DSR to provide more effective structural regularization for target-domain completion than either constraint alone.
Overall, these results demonstrate that all proposed modules are complementary, and their integration leads to substantial improvements in UDA point cloud completion.

\subsection{Visualization and Analysis}
\noindent \textbf{Visualization Results on 3D-FUTURE.}
Figure~\ref{3D-FUTURE-Visualization} presents qualitative comparisons on the 3D-FUTURE dataset. Compared with prior state-of-the-art methods, DAPointRWKV consistently produces more accurate and structurally complete reconstructions, with clear advantages in preserving both fine-grained local details and overall shape integrity. These improvements are particularly evident for categories such as Chair, Sofa, Lamp, and Table, as highlighted in the corresponding rows of the figure, where challenging missing regions are more faithfully recovered. These visual results further verify the robustness of DAPointRWKV in cross-domain point cloud completion.

\begin{table}[t!]
    \centering
    \caption{Computational efficiency of current state-of-the-art approaches: DAPoinTr, DAPointMamba, and DAPointRWKV.}
    \begin{tabular}{c|ccc}
    \toprule
         Model& Params(M)&FLOPs(G)&Time(ms) \\
         \midrule
         DAPoinTr~\cite{li2025dapointr}& 48.23 & 24.91&24.00 \\
         \midrule
         DAPointMamba~\cite{li2025dapointmamba}&36.77&5.19&3.96\\
         \midrule
         DAPointRWKV (Ours)&22.48&5.29&4.98\\
         \bottomrule
    \end{tabular}
    \label{table5}
\end{table}

\noindent \textbf{Computational Efficiency.} 
To evaluate the computational efficiency of DAPointRWKV, we compare it with the state-of-the-art DAPoinTr and DAPointMamba in terms of model parameters, FLOPs, and inference time. For fair comparison, all evaluations are conducted on an RTX 4090 GPU with 64GB RAM. We fix the batch size to 1 and the number of input points to 2048, and report the average inference time over 100 runs. As shown in Table~\ref{table5}, DAPointRWKV uses the fewest parameters among all compared methods, reducing the parameter count from 48.23M (DAPoinTr) and 36.77M (DAPointMamba) to 22.48M. In addition, DAPointRWKV is associated with only 5.29G FLOPs and 4.98 ms inference time, which are substantially lower than those of DAPoinTr (24.91G, 24.00 ms) and remain comparable to DAPointMamba (5.19G, 3.96 ms). Notably, while DAPointMamba and DAPointRWKV are both based on linear-complexity sequence modeling, our method achieves substantially better completion performance with comparable computational cost and inference latency. These results prove that DAPointRWKV achieves substantially better completion performance without sacrificing computational efficiency, yielding a more favorable efficiency-performance trade-off for cross-domain point cloud completion.

\noindent \textbf{Stable Analysis of DSR.}
We further evaluate DSR on 3D-FUTURE under different crop ratios (Table~\ref{table6}), corresponding to different re-observation strengths. DAPointRWKV obtains average CD values of 16.49, \textbf{16.02}, and 17.14 with 45\%, \textbf{55\%}, and 65\% crop ratios, respectively. Compared with GA-RWKV+RWKV-PD without DSR (CD:18.90), DSR consistently improves target-domain completion under different re-observation settings, further supporting its robustness and stabilizing effect under domain discrepancies. 
\begin{table}[t]
\centering
\footnotesize
\caption{Chamfer Distance (CD) on different crop-ratios.}
\label{table6}
\begin{tabular}{l|l|cccccc}
\toprule
\textbf{Crop Ratios}&\textbf{Avg}&\textbf{Cabinet}&\textbf{Chair}&\textbf{Lamp}&\textbf{Sofa}&\textbf{Table} \\
\midrule
45\%&16.49&19.77&15.48&14.36&19.34&13.49\\
\textbf{55\%}&\textbf{16.02}&\textbf{19.01}&\textbf{15.21}&\textbf{13.27}&\textbf{19.23}&\textbf{13.40}\\
65\%&17.14&19.39&15.98&14.05&21.19&15.08\\
\bottomrule
\end{tabular}
\end{table}

\section{Conclusion}
In this paper, we present Domain Adaptive RWKV (DAPointRWKV), a novel RWKV-based framework aimed at improving the transferability of RWKV for domain adaptive point cloud completion across diverse domains. Through the introduction of three tailored components, including the Geometry-Aware RWKV (GA-RWKV) Backbone, the RWKV-Oriented Progressive Decoder (RWKV-PD), and Dual Structural Regularization (DSR), DAPointRWKV effectively enhances local geometric modeling, bridges recurrent token representations with coarse-to-fine geometric reconstruction, and stabilizes unlabeled target-domain adaptation. Extensive experiments on both synthetic and real-world datasets demonstrate that the proposed DAPointRWKV consistently outperforms state-of-the-art methods, particularly in challenging categories.
\section{Acknowledgments}
Q. Zhou was supported in part by JSPS KAKENHI Grant Number JP26K21298.




\bibliographystyle{ACM-Reference-Format}
\balance
\bibliography{sample-base}










\end{document}